# A Feature Fusion-Net Using Deep Spatial Context Encoder and Nonstationary Joint Statistical Model for High Resolution SAR Image Classification

Wenkai Liang [1], Yan Wu [1,*], Ming Li [2], Peng Zhang [2], Yice Cao [1], Xin Hu [1]

[1] Remote Sensing Image Processing and Fusion Group, School of Electronic Engineering, Xidian University, Xi'an 710071, China
[2] National Key Laboratory of Radar Signal Processing, Xidian University, Xi'an 710071, China

**ABSTRACT:**

Convolutional neural networks (CNNs) have been applied to learn spatial features for high-resolution (HR) synthetic aperture radar (SAR) image classification. However, there has been little work on integrating the unique statistical distributions of SAR images which can reveal physical properties of terrain objects, into CNNs in a supervised feature learning framework. To address this problem, a novel end-to-end supervised classification method is proposed for HR SAR images by considering both spatial context and statistical features. First, to extract more effective spatial features from SAR images, a new deep spatial context encoder network (DSCEN) is proposed, which is a lightweight structure and can be effectively trained with a small number of samples. Meanwhile, to enhance the diversity of statistics, the nonstationary joint statistical model (NS-JSM) is adopted to form the global statistical features. Specifically, SAR images are transformed into the Gabor wavelet domain and the produced multi-subbands magnitudes and phases are modeled by the log-normal and uniform distribution. The covariance matrix is further utilized to capture the inter-scale and intra-scale nonstationary correlation between the statistical subbands and make the joint statistical features more compact and distinguishable. Considering complementary advantages, a feature fusion network (Fusion-Net) base on group compression and smooth normalization is constructed to embed the statistical features into the spatial features and optimize the fusion feature representation. As a result, our model can learn the discriminative features and improve the final classification performance. Experiments on four HR SAR images validate the superiority of the proposed method over other related algorithms.

## 1. INTRODUCTION

Synthetic aperture radar (SAR) imaging system has the ability to observe the earth surface without the constraints of illumination and cloud coverage (Moreira et al., 2013). It has become a very significant source of ground information in the field of modern remote sensing. SAR land cover classification is an important step in a variety of SAR image interpretations and applications, such as agricultural monitoring, urban planning, and damage assessment. (Rossi et al., 2015; M. Satake et al., 2013). With the development of new generation SAR sensors, e.g., TerraSAR-X (Breit et al., 2009), Gaofen-3 (Gu et al., 2015), and airborne SAR, large amounts of high-resolution (HR) SAR images have become available. Although the HR SAR image can provide sufficient detailed information of ground objects, it also presents more complex backscattering and spatial layout hard to deal with. Thus, pushing toward the novel HR SAR classification methods is urgently needed.

For the SAR image classification, effective feature extraction or feature learning is essential. In most cases, the discriminant ability of the features determines the quality of SAR image classification. The dominant discrimination of single-polarized SAR images is the amplitude or intensity information. Therefore, mining effective feature representations from the spatial context of a pixel plays a crucial role in the classification decision. According to the current research trend of feature extraction, the mainstream methods for SAR image classification can be roughly categorized as data-driven deep learning methods and scattering-based statistical analysis methods.

To depict the content of SAR images, most traditional methods rely on extracting intensity (Esch et al., 2011) and texture information (Dumitru et al., 2013). Some work also focuses on designing more discriminant handcrafted feature descriptors for SAR images, such as SAR histogram of oriented gradients (SAR-HOG) features (Song et al., 2016) and covariance of textural features (CoTF) (Guan et al., 2019). Compared to traditional methods, many studies have been proved the ability of deep neural networks (DNN) (Bengio et al., 2013) to automatically extract discriminative features of SAR images and achieve remarkable results with limited labeled data. Geng et al. (2018) proposed a deep supervised and contractive neural network (DSCNN) whose inputs are the combination of GLCM, Gabor, and HOG features for high-level feature learning. Zhao et al. (2017) proposed a discriminant DBN (DisDBN) for SAR image classification, which combined the ensemble learning with a deep belief network in an unsupervised manner. Chen et al. (2016) proposed an all-convolutional network (A-ConvNet) for SAR target recognition, which removes the fully connected layer and adds a dropout layer to prevent over-fitting. Fu et al. (2018) presented to use a deep residual network (ResNet) and introduced the dropout layer into the building block to alleviate overfitting caused by limited SAR data. A greedy hierarchical convolutional neural network (GHCNN) is developed to realize an efficient patch-based classification for single-polarized SAR images (Sun et al., 2020). In general, deepening the network can improve the expressive ability of features. However, training a very deep network can be difficult given the scarcity of SAR labeled data. Transfer learning becomes a common choice for applying deeper networks. Huang et al. (2021) proposed to transfer the pre-trained ResNet-18 model from NWPU-RESISC45 dataset to solve the large-scale HR SAR dataset classification. In (Zhang et al., 2021), a modified VGGNet (MVGG-Net) that transfers pre-training parameters from the ImageNet dataset is proposed for extracting deep features of SAR amplitude images. Besides, there are some attempts to use statistical methods to encode CNN features to improve the classification accuracy. Liu et al. (2020) proposed a statistical CNN (SCNN) for SAR land-cover classification, which characterizes the distributions of CNN features by the mean and variance statistics. Liang et al. (2021) integrated the covariance-based second-order statistics of CNN features, verifying that high-order statistics can improve the ability of CNN to distinguish various SAR land covers.

* Corresponding author
E-mail address: ywu@mail.xidian.edu.cn.

Due to the unique character of the coherent speckle, the statistical properties of SAR images also provide valuable information. The parametric approach is to postulate a given mathematical distribution for the statistical modeling, which has been intensively studied for SAR feature extraction due to simplicity and applicability (Li et al., 2011). Some non-Gaussian parametric models have been employed to extract statistical features of backscatters from different land covers, such as Rayleigh (Argenti et al., 2013), Gamma (Mart et al., 2014), K (Jen et al., 1984), Log-normal (LN) (Trunk et al., 1970), Weibull (Sekine et al., 1990), Fisher (Bombrun et al., 2008), and generalized Gamma (Li et al., 2011), etc. As the resolution of SAR images increases, the emergence of heterogeneous regions makes the modeling of HR SAR images is still challenging. To accurately describe the statistical properties of HR SAR images, two extended types of methods have been proposed. One is to use the basic probability model combination to build a stronger model to describe the SAR statistical properties. Many mixed statistical models, such as the Gamma mixture model (Nicolas et al., 2002), generalized Gamma mixture model (Li et al., 2016), and lognormal mixture model (Zhou et al., 2015) have been proposed. These models are then used in a Bayesian framework such as Markov random fields (Song et al., 2017) to implement classification. The other is to extend SAR images to complex value domains through transform domain methods such as Gabor transform (Lee et al., 1996), Wavelet transform (Zhange et al., 2012), and Contourlet transform (Golpardaz et al., 2020). By statistical modeling of multidimensional complex value subbands, the statistical features are more discriminant. Karine et al. (2017) used Weibull or Gamma distributions to model the dual-tree complex wavelet (DT-CWT) transform subbands of SAR images and stacked the statistical parameters obtained by each subband to form the statistical feature vectors. In (Karine et al. 2020), the author established the statistical dependence between DT-CWT subbands by introducing the Copula model (Sakji et al. 2009) and used the multivariate copula parameters constructed as the statistical features. Then, the above features are fed into a classifier such as the Softmax (Bridle et al. 1990) or the sparse representation (SC) (Wright et al. 2009) for classification.

There are a few works (Zhang et al., 2021; He et al., 2020; Ai et al., 2019) that try to fuse multiple features to improve the classification accuracy for SAR images. They focus on the fusion of deep features and other primary features such as polarization features and wavelet features. The effectiveness of statistical properties of the SAR image was ignored in these methods. Also, because each part of these fusion methods is individually learned, thus it cannot benefit from end-to-end learning. Reichstein et al. (2019) indicated that it is necessary to integrate the physical model and data-driven learning models in multiple ways to provide theoretical constraints when learning models from remote sensing data. As a result, the objective of this paper is to explore and design a framework that combines deep learning and statistical modeling to further improve the performance of the algorithm for SAR image classification.

To realize this goal, three important challenges remain. First, HR SAR images show complex structural and geometrical features, the above CNN models only use the single-scale convolution blocks that limit the scope of spatial information extraction. Transfer learning makes it easier for the deep model to learn the discriminative features from SAR images, but it ignores the inherent imaging mechanism differences between different datasets. In addition, traditional deep models such as VGGNet (Simonyan et al., 2014), and Resnet (He et al., 2016) contain a huge amount of parameters, which will increase the computational burden and memory consumption. More practical, the development of a more effective and efficient lightweight network is an inevitable requirement for intelligent SAR processing systems in the future. Second, through the statistical modeling of multiple wavelet sub-bands of the SAR image, and further establishing the dependence between the sub-bands, the expression ability of statistical features can be effectively improved. The common way is to use the coupla model to jointly model the dependence between multiple subband distributions. However, the coupla model usually needs to calculate a closed-form kullback-leibler divergence in the similarity measure. The parameter optimization process will be very time-consuming. Thus, another challenge is effectively designing a more efficient SAR global statistical feature representation scheme, and it can be effectively integrated with CNN features in linear space. Third, to fuse spatial and statistical features, the most direct way is to concatenate two types of features in a proportional weight parameter. However, the determination of the proportional parameter requires tedious experiments, which is difficult to put into practice. Hence, the third challenge is how to exploit the spatial and statistical information more effectively.

To address the aforementioned challenges, we propose a novel two-stream spatial-statistical feature extraction, feature fusion, and classification framework for HR SAR image classification. The proposed framework contains the following three modules: a spatial feature extraction module, a statistical feature extraction module, and a feature fusion module. At first, inspired by convolutional block attention module (Woo et al., 2018), group convolution (Huang et al., 2018), and dilated convolution (Yu et al., 2015), a new deep spatial context encoder network (DSCEN) is proposed to extract spatial features from SAR images with a small amount of labeled data. Second, inspired by the Marginal Distribution Covariance Model (MDCM) (Li et al., 2019), we introduce the covariance matrix to describe the multi-dimensional statistical properties of the wavelet subbands of the SAR image. This method can fully capture the high-order statistics of the SAR image and form a discriminant statistical feature descriptor. Finally, considering the complementarity of spatial and statistical features, a fusion network is proposed to fuse two types of features and the complete model is trained in an end-to-end manner. Compared with other SAR image classification methods, the proposed method can effectively combine the advantages of local spatial features and global statistical features, and the multi-feature information fusion in a unified training process can boost the robustness of the model for various land covers. The main contributions of this paper are listed as follows:

1. A new deep spatial context encoder network (DSCEN) with a lightweight structure is proposed to extract spatial features from SAR images. Our DSCEN consists of the multi-scale group convolution (MSGC) module and channel attention (CA) module. Specifically, the MSGC module can expand the scope of context information extraction in the spatial domain with few parameters. The CA module, located at the last layer of the network, is used to increase the interaction between high-level feature channels. Consequently, the proposed DSCEN is able to capture multi-scale contextual information in higher performance and can be effectively trained with the limited SAR labeled data, resulting in a more competitive spatial feature representation.

2. The nonstationary joint statistical model (NS-JSM) is first adopted to capture multidimensional scattering statistics in the

Gabor wavelet domain of SAR image and form a more distinguishable global statistical feature. Specifically, the NS-JSM uses different distributions to model the magnitudes and phases of Gabor wavelet subbands and then uses the covariance matrix to form the compact global statistical features for the mapped data in cumulative distribution function (CDF) space. The obtained statistical descriptor can not only capture statistical dependence and nonstationary correlation of SAR images that have not been explicitly considered by CNNs but also suppress the influence of noise.

3. A feature fusion network (Fusion-Net) base on group compression and smooth normalization is constructed to fuse spatial and statistical features and optimize the fusional feature representation. The feature Fusion-Net not only utilizes the complementary information of spatial and statistical features but also merges the statistical information into the network to participate in end-to-end training. Therefore, the feature representation ability and classification performance of the entire model are improved.

The rest of this article is organized as follows. Section 2 reviews some related works on CNN and the statistical distribution of SAR images. Section 3 first introduces the deep spatial feature extraction based DSCEN, then presents the statistical feature extraction based NS-JSM, and further proposes the feature Fusion-Net model. The experiments and results are presented in Section 4. Finally, some concluding remarks are drawn in Section 5.

## 2. RELATED WORK

### 2.1 CNN for SAR image classification

Compared with optical remote sensing images, single-polarization SAR images contain only intensity or amplitude spectrum. The traditional SAR image classification method is usually to extract the patch centred on the pixel to assist the classification of the central pixel. Thus, extracting effective spatial context information of a pixel plays a vital role in the decision of the pixel classes. Thanks to the efficiency with local connections, shared weights, and shift-invariance, CNNs have been applied on SAR images to exploit the spatial features.

Different from the optical remote sensing scene classification which directly takes the image as input, the pixels from the SAR image need to extract their local patches as the input of CNN to achieve pixel-level classification. Given a SAR image $(I_i)_{i \in N}$ with the size of $m \times n$, where $m$ and $n$ are the height and width of the spatial dimensions, respectively. $N = m \times n$ denotes the number of image pixels. $x_i, i = 1,...,N$ represents the gray value of the current pixel, $y_i, i = 1,...,N$ stands for the corresponding ground-truth label. First, all the labeled pixels together with their local patches $X_i$ with the size of $s \times s$ are extracted to form the samples. Then, training, validation, and test samples are constructed for training and evaluate the CNN model. The basic components of the CNN model contain a convolutional layer, nonlinear activation layer, and pooling layer. Generally, we treat a combination of a convolutional, activation, and pooling layer as a convolution block. Specifically, the major operations performed in the convolutional block can be represented as

$$F^l = pool\left(\sigma\left(F^{l-1} \otimes W^l + b^l\right)\right) \quad (1)$$

where $F^{l-1}$ is the input feature map of the $l$th layer, $W^l$ and $b^l$ are weights and bias of the $l$th layer, respectively. The input features $F^0$ of the first layer of CNN are the sample patch $X_i$. $\sigma(\cdot)$ denotes the activation function, which can be sigmoid, tanh, or leaky ReLU (lReLU) (Maas et al. 2013). $pool(\cdot)$ is the pooling operation for abstracting the feature maps.

The successive convolutional blocks are stacked together can extract the high-level features. Then, the output feature maps are flattened into a 1-D vector, and a fully connected classification is performed. In the end, the softmax function is connected on the last layer to form the class conditional probability distributions of each sample, which is defined as

$$p_i^c = \frac{e^{z_i}}{\sum_{t=1}^{T} e^{z_t}}, \quad c = 1,...,C \quad (2)$$

where $z_i$ is the vectorized feature vector of the last layer, $C$ is the total number of classes. To optimize the CNN model, the cross-entropy loss function (Abeyruwan et al. 2016) is adopted as the learning objective, which is defined as

$$L = -\sum_c c_i \log(p_i) \quad (3)$$

The mini-batch gradient descent algorithm is used to optimize the parameters of CNNs. After the model completed the training, the label of the test sample can be selected according to on the maximum probability as

$$\tilde{y}_i = \max_{c=1,...,C} p_i^c \quad (4)$$

Due to the limited available training samples of SAR images, some optimization trick modules can be used to speed up the training procedure and prevent overfitting. Generally, Batch-normalization (Ioffe et al. 2015) is connected to the convolutional layer to accelerate model convergence by preventing gradient vanishing. Data augmentation and dropout (Hinton et al. 2012) are used to prevent overfitting and further enhance network performance.

### 2.2 Statistical Models for SAR images

SAR is an active microwave imaging system that emits electromagnetic waves and receives the backscattered echo signals. Due to the coherent scattering processes at each pixel, the parametric scattering models of SAR images can be expressed as follows

$$X = X_{re} + jX_{im} = Ae^{i\theta} = \sum_{k=1}^{K} A_k e^{i\theta_k} \quad (5)$$

where $K$ is the number of discrete scatterers, $A_k$ and $\theta_k$ are the amplitude and phase of the $k$-th scatterer, respectively. $X_{re}$ and $X_{im}$ are the decomposition of $X$ in its real and imaginary parts. With the Gaussian assumption of $X_{re}$ and $X_{im}$, the amplitude $A$ follows a Rayleigh distribution, and the intensity $I = A^2$ has a negative exponential probability density function (pdf) (Argenti et al., 2013). In many practical cases, the statistical distribution of SAR images exhibits non-Gaussian behavior. Some prior hypothesis models such as Gamma, K distribution are proposed to fit the distribution of SAR data. Additionally, there are some empirical distribution models such as log-normal, Weibull, and Fisher distributions are obtained by experimental analysis on actual SAR images. Table 1 lists the PDF and CDF of some non-Gaussian statistical models.

**Table 1**
PDF and CDF of typical non-Gaussian statistical models for SAR image analysis.

| Model | PDF | CDF | Parameters |
|---|---|---|---|
| exponential | $f(r\|\sigma) = \theta e^{-\sigma r}, r \geq 0$ | $F(r) = 1 - e^{-\theta r}$ | $\theta$ |
| Rayleigh | $f(r\|\sigma) = \frac{r}{\sigma^2} e^{-r^2/(2\sigma^2)}, r \geq 0$ | $F(r) = 1 - e^{-r^2/(2\sigma^2)}$ | $\sigma$ |
| Gamma | $f(r\|\varphi,\beta) = \frac{\beta^\varphi r^{\varphi-1} e^{-\beta r}}{\Gamma(\varphi)}, r \geq 0$ | $F(r\|\varphi,\beta) = \frac{\gamma(\varphi, \beta r)}{\Gamma(\varphi)}$ | $\varphi > 0, \beta > 0$, |
| Log-Normal | $f(r\|\mu,\sigma) = \frac{1}{r} \frac{1}{\sigma\sqrt{2\pi}} \exp\left(-\frac{(\ln r - \mu)^2}{2\sigma^2}\right), r \geq 0$ | $F(r\|\mu,\sigma) = \frac{1}{\sigma\sqrt{2\pi}} \int_0^r \frac{1}{t} \exp\left\{\frac{-(\ln t - \mu)^2}{2\sigma^2}\right\} dt$ | $\mu, \sigma$ |
| Weibull | $f(r\|\xi,\nu) = \frac{\nu}{\xi} \left(\frac{r}{\xi}\right)^{\nu-1} e^{-(r/\xi)^\nu}, r \geq 0$ | $F(r\|\xi,\nu) = 1 - e^{-(r/\xi)^\nu}$ | $\xi > 0, \nu > 0$ |
| Nakagami | $f(r\|\varphi,\beta) = \frac{2\varphi^\varphi}{\Gamma(\varphi)\beta^\varphi} r^{2\beta-1} \exp\left(-\frac{\varphi}{\beta} r^2\right), r \geq 0$ | $F(r\|\varphi,\beta) = \gamma\left(\varphi, \frac{\varphi}{\beta} r^2\right)$ | $\varphi \geq 0.5, \beta > 0$ |

In SAR image segmentation or classification tasks, modeling only the amplitude or intensity of SAR images may not be enough for statistical features to have sufficient discriminant ability. To fully explore the statistical properties of the SAR image, a feasible way is to extend the SAR image to the wavelet domain to enhance the discriminative of the statistical features. In general, there is no specific statistical model for signals in the complex domain of SAR images. It is commonly accepted that the coefficients are highly non-Gaussian, exhibit heavy-tails (Kwitt et al., 2010). Note that parameter estimation is a key issue for the use of statistical models in SAR image processing. Accuracy of the model solution and complexity of the estimation has a great impact on the results and their usage (Li et al., 2011). Based on the above analysis, the combination of wavelet decomposition and simple pdfs mentioned above to capture the statistical properties of SAR images can not only improve the computational efficiency but also improve the discriminant ability, which may be a powerful and promising way.

## 3. THE PROPOSED METHODS

For effective HR SAR image classification, a novel end-to-end feature Fusion-Net framework is proposed to make full use of the complementarity among spatial and statistical information. The proposed framework is illustrated in Fig. 1, which consists of the following three steps: (1) spatial feature extraction using the deep spatial context encoder network (DSCEN). (2) statistical feature extraction using the nonstationary joint statistical model (NS-JSM). (3) spatial-statistical feature fusion and classification using Fusion-Net. Relevant details of each part are introduced in the following subsections.

### 3.1 Deep Spatial feature extraction

Contextual information, which reflects the underlying spatial dependencies between the central pixel and its surroundings, is pivotal to identify SAR ground objects. Some classical CNNs, such as VGGNet [45], and ResNet [46], exploit spatial context features by successive stacking standard convolution blocks. However, both the size of the input patch and the complexity of CNN should be considered when applying deep CNNs to the SAR image processing task. On the one hand, these networks contain too many pooling layers, which will overly contract the feature space of the small SAR patch, thus affecting the classification accuracy. On the other hand, the limited SAR labeled data is not enough to support the fully supervised very deep CNN training. In addition, the computational burden and memory consumption of a large CNN model are often faced with many practical limitations. Based on the above analysis, a new lightweight DSCEN model is designed for the SAR feature extraction. In DSCEN, the multi-scale group convolutional block and channel attention block are utilized to generate spatial feature representation for the SAR image. As shown in Fig. 1, it consists of two components: (1) multi-scale group convolutional (MSGC) block, and (2) channel attention block.

#### 3.1.1 Multi-Scale Group Convolutional Block

Generally, the deeper layers in CNNs contain the larger receptive field on the input image, which can learn more extensive spatial context features. Fig. 2(a) shows a standard convolutional block usually used in CNN, which is composed of two consecutive 3 × 3 convolutional layers. There are two imperfections that need attention. One is that a single standard convolution block has a constant receptive field, so it cannot extract multi-scale spatial structure information. The other is that as the number of feature channels increases, it will lead to a heavy-weight model. A direct way to stacks several standard convolutional blocks to enlarge the field of view. However, this will increase the complexity of the model. Also, Kwitt et al., (2010) points out that the actual receptive field of a single convolution block is much smaller than the theoretical size.

To solve the above problem, we propose an MSGC block to take full advantage of local and surrounding context features. The dilate convolution and group convolution are applied to the MSGC block to make our models more effective and efficient. Fig. 2(b) presents the structure of the MSGC block. The top branch of the block is a standard 3 × 3 convolutional layer, which is used to perform dimension transformation and increase the nonlinearity. The bottom branch contains two different types of feature extractors to mine spatial information at different sizes. Among them, the 3 × 3 G-Conv layer is used to learn local detailed features, and the 3 × 3 DG-Conv layer is used to capture broader context features. Fig. 3 gives a comparison of the standard convolution, G-Conv and DG-Conv. From the top of Fig. 3, we can see the difference between the standard convolution kernel and the dilate convolution kernel. The main idea of dilated convolution is to insert zeros in convolutional kernels to increase the receptive field while reducing the para-meters. As shown at the bottom in Fig. 3, group convolution divides all $R$ channel inputs into $G$ groups, and each group corresponds to $R/G$ channels. In group convolution, there are no connections across the channels in different partitions, and it can be regarded as the structured sparse. Compared with standard convolution, group convolution can reduce the number of model parameters and may also achieve better performance. Besides, we further compress the number of feature channels to half of the input in both G-Conv and DG-Conv layers. A simple concatenation layer followed by the BN and lReLU operators is utilized to aggregate multi-scale features. Finally, the proposed MSCB block can achieve a significant performance boost while keeping the model complexity to a minimum.

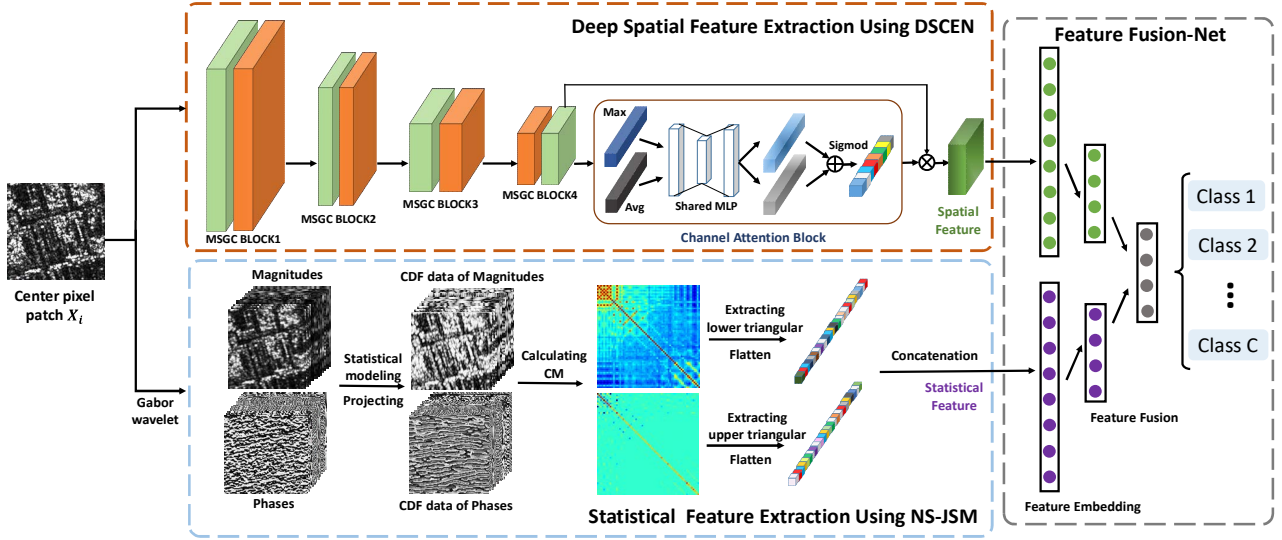

**Fig. 1.** Framework of the proposed feature Fusion-Net for SAR image classification.

### 3.1.2 Channel Attention Block

HR SAR images contain complex structural and geometrical information, which may lead to high intra-class variance and low inter-class difference. The commonly used CNNs consider that each channel feature maps in one layer have the same importance. Thus, it may not be able to highlight the importance of some salient feature maps, failing to distinguish complex scenes. Inspired by the attention mechanism (Woo et al., 2018), our goal is to enhance the feature representation power by channel attention (CA) block. The intuitive way is to add a CA block behind each convolution layer. However, it is expensive to perform an independent CA block on each convolutional layer. Based on insights about CNN properties from (Zeiler et al., 2014), low-level features that were close to input extracts more local spatial information, and high-level features that were close to the classifier encode more semantic information. To this end, we apply the CA block in the last layer of DSCEN, which can pay more attention to the meaningful class-specific information for the current task efficiently.

In the CA block, the global average and global max pooling operators are applied to aggregate spatial information of the input features. Then, two feature descriptors are fed into a shared multilayer perceptron (MLP) with one hidden layer (where the number of the hidden layer units is $FN/re$, $re$ is the reduction ratio) to capture the channel-wise dependencies. Finally, we merge the output feature vectors by using an element-wise addition and a sigmoid function. The obtained channel attention vector can be computed as

$$M_c(\mathrm{F}) = sigm\big(MLP(AvgPool(\mathrm{F})) + MLP(MaxPool(\mathrm{F}))\big) \quad (6)$$

where $sigm$ denotes the sigmoid function, $MLP$ is a shared multilayer perceptron. The flowchart is shown in the channel attention block of Fig. 1. Finally, we can obtain the more discriminative output features by employing a scale layer to re-weight the high-level features with the channel attention vector. It should also be noted that through end-to-end training, the network is capable of adaptive learning the weights of the feature maps, thereby focusing on important features and suppressing useless ones more efficiently.

### 3.1.3 Network Architecture

Based on the MSGC block and CA block, we construct the DSCEN model to extract the spatial features from the SAR image patch. As shown in Fig.1, the proposed DSCEN model contains four MSGC blocks and one CA block. The number of output channels of each MSGC block is 16, 32, 64, and 128, respectively. At the end of each MSGC block, a max-pooling layer is followed to abstract the feature maps. In addition, dropout is added after each pooling layer to prevent overfitting. Given an input image patch $X_i$ centered on the pixel $x_i$, whose spatial feature $\mathrm{F}^{Spatial}$ can be conducted multiple MSGC block-based convolutions and one CA block-based recalibration in DSCEN. The formula for calculating the spatial feature $\mathrm{F}^{Spatial}$ is described as follows

$$\mathrm{F}^{Spatial} = M_c\big[MSGC_4\big(MSGC_3\big(MSGC_2\big(MSGC_1(X_i)\big)\big)\big)\big] \quad (7)$$

where $MSGC$ represents a multi-scale group convolution operation. When the spatial feature representation is obtained, it can be imported to the feature fusion network for classification.

### 3.2 Statistical feature extraction

High-order scattering statistics of the SAR image provide much valuable information for data representation. However, CNN-based methods rarely consider and exploit the statistical properties of the SAR image. It is necessary to merge the statistical information into CNN to improve the feature representation. Due to the limitation of the lack of spectrum, we extend the single-polarization HR SAR image to the Gabor wavelet domain to model it statistically and make the statistical feature more discriminative.

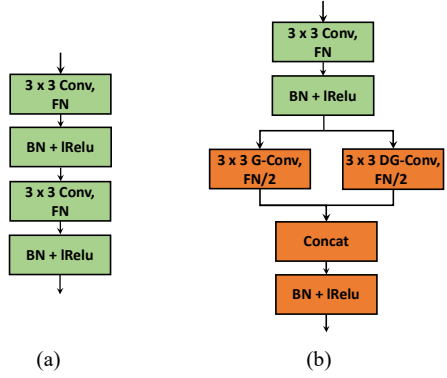

**Fig. 2.** Illustration of (a) Standard convolutional block. (b) Multi-scale group convolutional block. ("G-Conv" represents the group convolution, "DG-Conv" represents the dilate group convolution, "FN" represents number of feature channels)

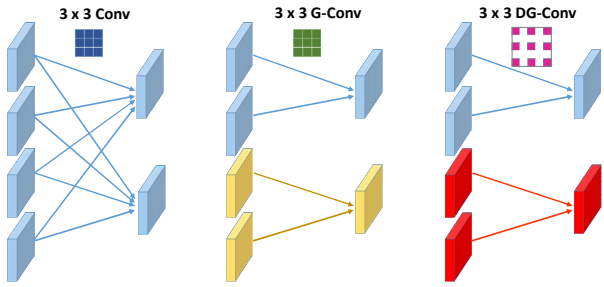

**Fig. 3.** Comparison of (a) Standard convolution. (b) Group convolution. (c) Dilate group convolution.

Considering that the Gabor filter has scale and direction selection characteristics, it is compatible with the direction sensitivity of SAR images. Therefore, the Gabor filter is chosen to transform the SAR image into the complex wavelet domain. Given a SAR image $I$ and the Gabor filter $G_{u,v}$, the Gabor wavelet subbands can be computed as follows

$$O_{u,v}(z) = I(z) * G_{u,v}(z) \tag{8}$$

where $*$ is the convolution operator, $z = (z1, z2)$ denotes the coordinates in the spatial domain. $v = 1,...,V$ and $u = 1,...,U$ represent the scale and direction. $U$ and $V$ are the number of scales and directions of Gabor filters, respectively. $O_{u,v}(z)$ is the complex number result. Due to the phase of the wavelet is also important discriminant information, we extract the magnitude and phase of the Gabor wavelet simultaneously. Then, the magnitude $M_{u,v}(z)$ and the phase $P_{u,v}(z)$ of the Gabor filter output is computed as

$$M_{u,v}(z) = \sqrt{(Re(O_{u,v}(z)))^2 + (Im(O_{u,v}(z)))^2} \tag{9}$$

$$P_{u,v}(z) = \arctan(Im(O_{u,v}(z))/Re(O_{u,v}(z))) \tag{10}$$

For the convenience of subsequent representation, the magnitude and phase subbands are vectorized and organized into the observation matrix $M_{mag} \in \mathbb{R}^{n \times d}$ and $M_{pha} \in \mathbb{R}^{n \times d}$, respectively. Each column of $M_{mag}$ or $M_{pha}$ is the observations of the magnitude variable $m_j$ or phase variable $a_j$, $j = 1,...,d$. Here, each column corresponds to a subband. $n$ is the total number of the variable in a subband, and $d = U \times V$ represents the number of subbands.

For the observations of a subband, it is usually to make an underlying assumption based on a specific distribution and then compute the distribution parameters to build the feature vectors. Supposing the observations in $j$-th column of $M_{mag}$ or $M_{pha}$ obey a certain marginal distribution which CDF is $F_j(r|\theta)$ and PDF is $f_j(r|\theta)$. Then, we can use the observations of variables to estimate the distribution $f_j(r_j|\theta_j)$ and parameter $\tilde{\theta}_j$ by maximum likelihood or moments. Yu et al. (2010) concluded that non-Gaussian pdfs are suitable to model the statistical behavior of the magnitude of Gabor wavelet subbands. Considering simplicity and applicability, the log-normal distribution was used to fit the distribution of the magnitude subbands, and the uniform distribution is adopted to model the distribution of the phase subbands in our work. In addition, the effects of the different non-Gaussian distributions mentioned in Section 2 to model the Gabor magnitude subbands are validated in the experiments.

Afterward, inspired by (Li et al., 2019), the observations of each column are projected to its corresponding CDF space by using CDF $F_j(r_j|\tilde{\theta}_j)$. The CDF space of the amplitude subband can be expressed as $M_F \in \mathbb{R}^{n \times d}$

$$M_F = [F_1, F_2, \cdots, F_d] = \begin{bmatrix} F_{1,1} & F_{1,2} & \cdots & F_{1,d} \\ F_{2,1} & F_{2,2} & \cdots & F_{2,d} \\ & & \ddots & \\ F_{n,1} & F_{n,2} & & F_{n,d} \end{bmatrix} \tag{11}$$

Where, $F_i = [F_{1,i}, F_{2,i}, \cdots, F_{n,i}]$ is the detailed CDF vector for each column.

The most direct way is to construct the statistical descriptor by utilizing whole statistics. However, this strategy will lead to a very high-dimensional vector while ignoring the statistical dependence and nonstationary characteristics between subbands. To overcome this limit, the NS-JSM is used to describe these subbands statistics, which can form a more compact and robust statistical descriptor. To conveniently calculate the covariance matrix (CM), we change $M_F$ into the following form, $\tilde{M}_F = (M_F)^T = [Z_1, Z_2, \cdots, Z_n]$, where $Z_l = [F_{l,1}, F_{l,2}, \cdots, F_{l,d}]^T$, $l = 1,...,n$. Finally, the magnitude-based statistical feature can be represented by $d \times d$ CM of the subbands statistics

$$C_{mag} = \frac{1}{n-1} \sum_{l=1}^{n} (Z_l - \mu)(Z_l - \mu)^T \tag{12}$$

Where, $\mu$ is the mean of the feature vectors $Z_l$, $l = 1,...,d$. Through the above calculation method, we can also obtain the CM $C_{pha}$ corresponding to the phase-based statistical feature.

There are two advantages of the NS-JSM: 1) Covariance matrix builds the dependence and nonstationarity between two different subbands, which can fuse complementary information coming from different subbands and form the more compact and discriminative feature. 2) there is an average filter during covariance computation, which can further reduce the effect of speckle and noise. To form the final joint statistical descriptor based on magnitude and phase, we flatten and concatenate the two types of covariance features. Since the CM here is symmetric, we only use the upper triangular part of the magnitude CM and the lower triangular part of phase CM. Notably, the CM usually resides on the Riemannian manifold of the SPD matrix (Arsigny et al., 2007). Logarithmic transformation is applied in our work to map the covariance matrix to the linear space, and the final statistical feature descriptor of the SAR image patch can be expressed as

$$F^{Statistical} = triu\left(\log\left(C_{mag}\right)\right) \| tril\left(\log\left(C_{pha}\right)\right) \quad (13)$$

where $\|$ denotes the operation of concatenating. The scheme of the global statistical feature extraction process is shown in Fig.1.

### 3.3 Spatial-statistical feature Fusion-Net

After obtaining the spatial and statistical features, it is important to effectively fuse them for classification. A commonly used method is to utilize a weighted strategy to fuse the two types of features. However, it required a large number of experiments to find the optimal weight parameter. Also, this strategy cannot benefit from end-to-end learning to obtain more robust performance. To utilize the complementary information between the two types of features, we propose to train a two-layer perceptron feature fusion network (Fusion-Net) that can embed the statistical features into the spatial features in nonlinear feature space. In Fusion-Net, we use a sparsely connected network based on group convolution in the first layer for dimension reduction on the two features. Then, we adopt a fully connected network in the second layer to further fuse and optimize features so as to enhance feature discrimination. We define the fusion scheme as follows

$$F^{Fuse} = sigm\left(W_2\left(sigm\left(GConv(W_1', F^{Spatial})\right)\right) \\ \| sigm\left(GConv(W_1'', F^{Statistical})\right)\right) \quad (14)$$

where $W_1'$ and $W_1''$ are the weights of the first sparsely connected layer, respectively. $W_2$ is the weights of the second fully connected layer. $GConv$ denotes the group convolution. The proposed Fusion-Net has the following advantages: 1) Due to spatial and statistical features have high dimensions, the sparsely connected layer can achieve dimensional reduction with fewer parameters. It can also suppress useless information in each feature before fusion. 2) The sigmoid function as a smooth normalization mechanism can transform the features into a relatively consistent space. This choice can prevent either feature to become dominant, thus encouraging the contribution from both features. 3) Compared to the fully connected network, Fusion-Net has fewer parameters to learn complementary information, so it can further prevent overfitting.

Finally, the fusional feature vectors directly input to the softmax function to generate the predicted labels. In the training stage, the cross-entropy loss is adopted as the objective function. The parameters of the proposed method are trained in an end-to-end manner through iterative methods. Thus, the CNN-based spatial information and the global statistical information can interact during the training process, and the classification performance can be significantly improved.

## 4. EXPERIMENT

In this section, we evaluate the performance of the proposed method for HR SAR image classification. The data set description, detailed experimental setup, and experimental results with reasonable analysis are presented below.

### 4.1 Description of the Datasets

Four real HR SAR images obtained from different sensors were used to validate the effectiveness of the proposed method. These four data are high resolution and contain complicated structural and geometrical scene information. For each dataset, the ground truth images are generated by manual annotation according to the associated optical image, which can be found in Google Earth. The first HR SAR image was acquired by the TerraSAR-X satellite over the scene of Lillestroem, Norway, in 2013. It has 2675 ×1450 pixels in size with an HH-polar imaging mode. The acquisition mode of the data is staring spotlight and the resolution of the image is 0.5m. This scene contains five kinds of ground objects: Water, residential, roads, woodland, and open land. The original image and the ground truth image are shown in Fig. 8(a) and Fig. 8(b). The dataset is available at http://www.intelligence-airbusds.com.

The second data set was collected by the Chinese Gaofen-3 satellite in Guangdong Province, China, in 2017. The image size is 2600 ×4500 pixels, and the spatial resolution is 0.75 m. The image is HH-polarization data with the sliding spotlight mode. This data consists of seven classes: Mountains, water, building, roads, woodland, and open land. The original image and the ground truth are shown in Fig. 9(a) and Fig. 9(b).

The third data was from the area of Shaanxi province, China, collected by a Chinese airborne sensor in 2016. It was provided by the China Electronics Technology Group Corporation (CETC) Institute. This data has a size of 1800 × 3000 with a spatial resolution of 0.3 m. The image is HH-polarization data with spotlight mode. Seven classes are considered for the experiment: open land, roads, rivers, runway, woodland, residential, commercial. The original image and the ground truth image are shown in Fig. 10(a) and Fig. 10(b).

The fourth data was acquired by an X-band F-SAR sensor of the German Aerospace Center in 1989. The data set was from the Bavaria region in Germany, whose spatial resolution is 0.67 m. It contains 6187 ×4278 pixels and the polarization of the data is VV-polar mode. Four classes of interests were considered: water, residential, vegetation, and open land. The original image and the ground truth image are shown in Fig. 11(a) and Fig. 11(b), respectively. The dataset is available at https://www.dlr.de.

### 4.2 Experimental setup and evaluation metrics

The proposed classification method consists of three parts: spatial feature extraction using DSCEN, statistical feature extraction using NS-JSM, spatial-statistical feature fusion, and classification using Fusion-Net. First, we set the public parameters and training strategies involved in model training. Then, the detailed parameter determination for each module is discussed in the subsequent part. During the training phase, the unique loss function is optimized by the Adam optimizer (Kingma et al., 2015) with a constant learning rate of 0.001. The mini-batch size is set to 100 and the number of epochs is set as 150. In our DSCEN, all the convolutional weights are initialized from Gaussian distributions with zero mean and a standard deviation of 0.01, and no bias term. Moreover, the dropout ratio is set to 0.2.

To achieve pixel-based classification, training, validation, and test samples are necessary to be collected. In our experiment, all the labeled pixels together with their neighborhood patches are extracted to form the samples. The patch spatial size of 64 × 64 pixels was chosen as inputs. The input data is normalized in the range of 0-1 by max-min normalization. For each category, three hundred samples were randomly selected and divided into training and validation, accounting for 70% and 30%. Two forms of data augmentation including rotation and flipping were applied for training samples, which can increase the number of training samples to eight times the original.

In the testing phase, the network weights of the minimum loss on the validation data were loaded for evaluating the test data. A stride greater than 1 was used to infer the test data to avoid excessive computational costs. The stride was set to 5 in our work. The obtained class probability map was then upsampled the original resolution with a negligible loss in the accuracy. Notably, the parameters determination and analysis of the TerraSAR-X SAR image are discussed in the subsequent ablation study. The same parameter settings were used to classify the other three images. The specific trend analysis of the other three HR SAR images is the same as the TerraSAR-X SAR image. Here, we hope to avoid parameter tuning for each dataset and apply the optimization model to other datasets. It can more effectively verify the generalization performance of the model while reducing time consumption for actual application scenarios.

To reduce the influence of random initialization, each experiment was run five times independently. The overall accuracy (OA), average accuracy (AA), and Kappa coefficient (κ), and class-specific accuracy are used for evaluating the classification results. All experiments were conducted by MATLAB 2014a on the platform of a computer with I7 3.2-GHz CPU and 32-GB memory. The whole deep CNN system was built by the MatConvNet library.

**4.3 Analysis of DSCEN**

In the proposed DSCEN, the feature number of the network, the depth of the network, the effect of the MSGC block, and the effect of the CA block were discussed in detail as follows. When analyzing the effect of a particular block, we vary this parameter whilst fixing all others.

**4.3.1 Effect of feature number of network**

In general, the feature number in CNN can determine the diversity of features. Too few feature channels in CNN may not express sufficient discriminability, and too many feature channels introduce overfitting and increase model complexity. First, the DSCEN depth is fixed as 4. Then, four networks were defined to test the effect of feature number by reducing or adding parameters of the convolutional layers. We follow roughly the rule that the feature number in a convolutional block is twice that of the previous block. These classification results are displayed in Table II. From this table, we can observe that model setting 16-32-64-128 obtained the best performance. The DSCEN with fewer parameters is limited on classification performance, while DSCEN with more parameters may not be sufficiently trained due to limited training samples. Therefore, the model setting 16-32-64-128 is select as the default setting of the DSCEN in our experiment.

**Table 2**
Classification accuracy of different feature number.

| Feature number | OA | AA | κ |
|---|---|---|---|
| 8-16-32-64 | 0.8786 | 0.8812 | 0.8312 |
| **16-32-64-128** | **0.8900** | **0.8872** | **0.8462** |
| 32-64-128-256 | 0.8811 | 0.8842 | 0.8345 |
| 64-128-256-512 | 0.8808 | 0.8785 | 0.8338 |

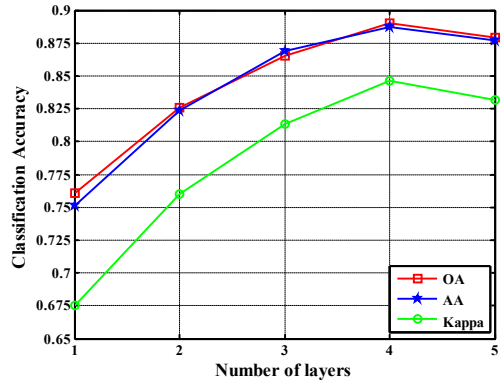

**Fig. 4.** The classification accuracy of the DSCEN with different layers.

**4.3.2 Effect of depth of DSCEN**

Deepen the network layers can extract the more abstract feature and improve classification performance. In DSCEN, the number of MSGC blocks determines the depth of the network. Since there is a pooling layer behind each MSGC block, the receptive field of DSCEN will become larger as the number of layers deepens, and the size of the feature map will be further compressed. Thus, we conducted experiments to test the effect of different depths by reducing or adding MSGC blocks. Fig. 4 shows the classification accuracy produced by the DSCEN with different layers. It can be seen that the accuracy tends to increase gradually as the network deepens. When the network deepens to the fifth layer, the accuracy begins to decay. This is because the SAR image patch size is small, the feature space is overly contracted, and causes too much resolution loss, thus affecting the classification accuracy. The best performance can be achieved by setting the network depth to four, which we use as the default setting for the DSCEN in our experiments.

**4.3.3 Effect of MSGC block**

There are three important components in the MSGC block that need to be compared, namely multi-scale convolution dilate convolution, and group convolution. Therefore, we define seven convolutional blocks to evaluate the effectiveness of the MSGC block. Specifically, we fixed the network at four layers and only used different convolutional blocks instead of the MSGC block for comparison. We used M-1 to indicate that only a standard 5 × 5 convolutional layer adopted in the block. The M-2 was used to denote that two standard 3 × 3 convolution layers were applied. We use a 3 × 3 compressed convolution layer and a 5 × 5 compressed convolution layer to replace the bottom convolution block in MSGC block, which is defined as M-3. The main difference between it and the MSGC block is that no dilate convolution and group convolution.is used. Further, a 3 × 3 dilate convolutional layer replaces the 5 × 5 convolution layer in M-3 as M-4. It is similar to the MSGC block, but here the grouping coefficient $G$ is set to 1. M-5 to M-7 were constructed mainly to verify the performance of group convolution. The main difference is that they used grouping coefficients $G$ of 2, 4, and 8 in block, respectively. The experimental results are shown in Fig. 4.

**Table 3**
Classification accuracy of different convolutional block.

| Model | Size | OA | AA | κ |
|---|---|---|---|---|
| M-1 | 1115K | 0.8596 | 0.8633 | 0.8057 |
| M-2 | 1311K | 0.8683 | 0.8789 | 0.8177 |
| M-3 | 1857K | 0.8845 | 0.8827 | 0.8388 |
| M-4 | 1177K | 0.8817 | 0.8815 | 0.8351 |
| M-5 | 682K | 0.8858 | 0.8866 | 0.8406 |
| M-6 | 650K | **0.8900** | **0.8872** | **0.8462** |
| M-7 | 634K | 0.8890 | 0.8867 | 0.8449 |

**Table 4**
Classification accuracy of different reduction ratio $re$ in CA block.

| CA block | OA | AA | κ |
|---|---|---|---|
| No-CA | 0.8823 | 0.8814 | 0.8359 |
| CA($re=1$) | 0.8898 | 0.8866 | 0.8459 |
| CA($re=4$) | **0.8900** | **0.8872** | **0.8462** |
| CA($re=16$) | 0.8858 | 0.8850 | 0.8407 |
| CA($re=32$) | 0.8826 | 0.8846 | 0.8366 |
| CA($re=64$) | 0.8875 | 0.8859 | 0.8428 |
| CA($re=128$) | 0.8883 | 0.8867 | 0.8439 |

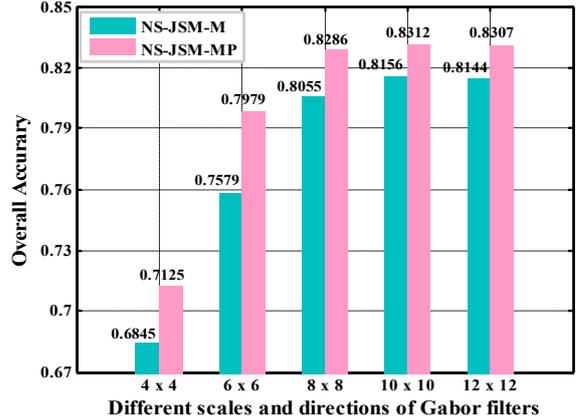

**Fig. 5.** Effect of NS-JSM with different settings on the overall accuracy.

**Table 5**
Classification accuracy of different statistical models.

| Statistical models | Type | OA | AA | κ |
|---|---|---|---|---|
| No | Mag | 0.7644 | 0.7775 | 0.6806 |
|  | Mag-Pha | 0.8171 | 0.8220 | 0.7532 |
| Exponential | Mag | 0.8055 | 0.8202 | 0.7374 |
|  | Mag-Pha | 0.8271 | 0.8339 | 0.7623 |
| Log-normal | Mag | 0.8086 | 0.8240 | 0.7389 |
|  | Mag-Pha | **0.8286** | **0.8354** | **0.7643** |
| Nakagami | Mag | 0.8063 | 0.8210 | 0.7359 |
|  | Mag-Pha | 0.8280 | 0.8346 | 0.7635 |
| Rayleigh | Mag | 0.8067 | 0.8198 | 0.7362 |
|  | Mag-Pha | 0.8287 | 0.8348 | 0.7644 |
| Weibull | Mag | 0.8069 | 0.8226 | 0.7366 |
|  | Mag-Pha | 0.8282 | 0.8350 | 0.7636 |
| Gamma | Mag | 0.8078 | 0.8232 | 0.7377 |
|  | Mag-Pha | 0.8285 | 0.8353 | 0.7641 |

From Fig. 4, we can see that the M-6 MSGC block with grouping coefficient $G=4$ can obtain the best classification results. Compared with the standard M-1 and M-2 blocks, we can conclude that introduction of multi-scale convolution can achieve better performance, mainly because it expands the receptive field and captures more contextual information. Then, comparing the M-3 and M-4 models, it can be observed that the introduction of dilate convolution can basically keep the accuracy unchanged while reducing the number of parameters. In models M-5 to M-7, the parameter amount is further reduced and the classification performance is improved by introducing group convolution. This is because the sparse property of group convolution may reduce feature redundancy and improve the feature learning ability of the MSGC block. In summary, we use the M-6 module as the default setting of DSCEN. It can significantly improve the classification results whilst reducing the number of model parameters.

### 4.3.4 Effect of CA block

To assess the effectiveness of the CA block, we compare the performance of DSCEN without the CA block to the version with the CA block for different reduction ratios $re \in \{1, 4, 16, 32, 64, 128\}$. The experimental results are reported in Table 4. Intuitively, we can see that adding the CA block can produce better accuracy than without CA blocks in DSCEN. This explicitly demonstrates that the CA block can boost the classification performance of our model. Moreover, the results in Table 4 show that the accuracy does not gradually increase as $re$ decreases. The reason is that excessive compression of global feature descriptors may not capture better feature channel-wise interaction relationships. For this experiment, the CA block with a reduction ratio $re=4$ can achieve the best performance, and we select it as the default setting for subsequent experiments

### 4.4 Analysis of NS-JSM

#### 4.4.1 Effect of modeling of the Gabor wavelet subbands

We tested the effect of statistical modeling of the Gabor wavelet subbands on the classification results. This section has different degrees of freedom such as the choice of scales and directions of Gabor filters and whether to use phase features. Here, we varied the scales and directions of Gabor filters to $4 \times 4$, $6 \times 6$, $8 \times 8$, $10 \times 10$, and $12 \times 12$ to observe the classification results. Meanwhile, we used the symbols "NS-JSM-M" to indicate that only the magnitude (where Log-Normal distribution is used for magnitude modeling) of the Gabor wavelet subband is constructed for statistical features. Similarly, the symbol "NS-JSM-MP" represents that the magnitude and phase of Gabor subbands are used to construct the statistical features by the NS-JSM. Fig. 5 reports a comparison of different model settings on OA. As shown in Fig. 5, the increasing of scales and directions generally leads to an improvement in the classification accuracy. It can be observed that stable accuracy appears when the number of scales and directions is set to 8. For more scales and directions, there will be almost no improvement in accuracy. Furthermore, we can see that combining phase information can further improve the recognition rate of statistical features, which has been ignored in many wavelet-based classification tasks.

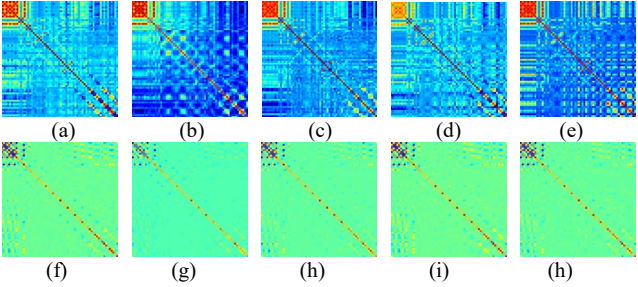

**Fig. 6.** The proposed covariance matrix for different classes of the SAR image. The upper row is the CM based on the amplitude modeling of the Gabor subbands; the lower row is the CM based on the phase modeling of the Gabor subbands. (a)-(f) Waters. (b)-(g) Residential. (c)-(h) Wood land. (d)-(i) Open land. (e)-(h) Road.

**Table 6**
Classification accuracy of different hidden units in Fusion-Net.

| Hidden units | OA | AA | κ |
|---|---|---|---|
| 32 | 0.9004 | 0.9047 | 0.8610 |
| 64 | 0.9078 | 0.9032 | 0.8704 |
| 128 | **0.9095** | **0.9055** | **0.8727** |
| 256 | 0.9027 | 0.9039 | 0.8639 |
| 512 | 0.9060 | 0.9025 | 0.8680 |

#### 4.4.2 Effect of different statistical models

To verify the fitting ability of the statistical modeling, different statistical models mentioned in Section 2 were used to model the magnitude subbands of Gabor wavelets. Also, the uniform distribution is adopted to model the phase subbands. The comparison results are presented in Table 5, where we use the symbols "Mag" and "Mag-Pha" to represent only the amplitude modeling of the gabor subband and the amplitude and phase modeling of the gabor subband at the same time to construct statistical feature. From Table 5, we observed that by modeling the Gabor subbands, NS-JSM has an improvement of about 1.5% on accuracy compared to the non-modeling solution. This indicate that by projecting the sub-bands coefficients into the CDF space, the noise coefficients in the sub-band may be smoothed, so that the more robust performance can be obtained. In addition, we can see that phase information provide a very important contribution in this method. Further, the difference in accuracy of different statistical models is almost small, which may be because these non-Gaussian distribution models are too simple to better model different types of image patches. But these models can solve the parameters faster. A more complex statistical model such as the mixed statistical model is another choice, but parameter estimation prevents it from being efficiently applied to the NS-JSM.

#### 4.4.3 Feature Visualization and Analysis

To illustrate the discriminative performance of the NS-JSM, we present the amplitude and phase covariance matrices corresponding to five different types of image patches from TerraSAR-X images in Fig. 6. It can be seen that the covariance features of different categories have different manifestations. Therefore, to a certain extent, the use of global statistical covariance features can distinguish the category attributes of different image patches.

### 4.5 Analysis of Fusion-Net

#### 4.5.1 Effect of node number and activation of Fusion-Net

To evaluate the performances of the proposed Fusion-Net, we conduct experiments to test the effect of different node numbers embedded in the neural network. To simplify the experiment, the number of nodes in the sparsely connected layer of the first layer and the number of nodes in the fully connected layer of the second layer in Fusion-Net are set to be equal. Then, we selected the number of nodes as 32, 64, 128, 256, and 512 for comparison. The results in Table 6 show that setting the number of nodes to 128 can produce the best classification results. When the number of nodes is reduced, the feature representation ability of the converter and the fusion layer will be weakened. As the number of nodes increases, the accuracy does not increase further, but it increases the number of parameters. Therefore, considering the model complexity and classification accuracy, we set the number of nodes is set to 128 in our experiment.

In addition, to illustrate the effectiveness of smooth normalization, we test the classification performance of the sigmoid and ReLU activation in Fusion-Net. Table 7 shows the comparison of our experiments. It is clearly seen that sigmoid activation yield better classification results than ReLU activation. The possible reason is that the feature value range of ReLU is from zero to infinite. Thus, there may be some outliers that affect the performance of the fusional feature. The sigmoid activation, as a normalization mechanism, can transform the characteristics into a relatively consistent space with a range of zero to one. It can maintain more detailed information, so that the Fusion-Net can mine complementary information more effectively, thus improving classification accuracy.

**Table 7**
Classification accuracy of different activation in Fusion-Net.

| Activation | OA | AA | κ |
|---|---|---|---|
| Sigmoid | **0.9095** | **0.9055** | **0.8727** |
| ReLU | 0.9038 | 0.9036 | 0.8651 |

**Table 8**
Comparison of different fusion schemes on classification accuracy.

| Schemes | OA | AA | κ |
|---|---|---|---|
| Prob-Fusion | 0.8994 | 0.8969 | 0.8590 |
| Feature-Fusion | 0.8955 | 0.8998 | 0.8541 |
| Fusion-Net | **0.9095** | **0.9055** | **0.8727** |

#### 4.5.2 Comparison of different fusion schemes.

To validate the proposed feature fusion scheme, we compare it with two other empirical fusion schemes including probability fusion and feature fusion. For probability fusion, we adopt the weight fusion to fuse the class probability maps of the DSCEN and the NS-JSM. The optimal weight is selected by trial and error. For feature fusion, we directly concatenate the output features of the DSCEN and the NS-JSM. Then, the fusion feature is fed into the softmax classifier for classification. Table 8 shows the comparison of our experiments. From Table 8, we can see that our Fusion-Net has about 1% improvement in OA compared to the empirical fusion method. The results indicate that the proposed fusion scheme can effectively fuse the spatial and statistical features, and benefit from the end-to-end training manner, the prediction accuracy of the entire model can be further refined.

#### 4.5.3 Feature Visualization

To further illustrate the ability of our Fusion-Net to extract complementary information between spatial and statistical features, Fig. 7 visualizes the distribution of three different features, which are spatial, statistical features, and fusion features, respectively. The distribution is generated by using the t-distributed stochastic neighbor embedding (t-SNE) (Van et al., 2008) algorithm on 1000 patches (200 patches per category) randomly selected from the test data. It can be observed that the feature distribution of our fusion network has fewer overlaps than the other two types of features. For the output features of the Fusion-Net, the distance between different classes of features becomes larger. The results show that our Fusion-Net has the capacity to achieve even superior feature recognition compared with the single spatial or statistical feature.

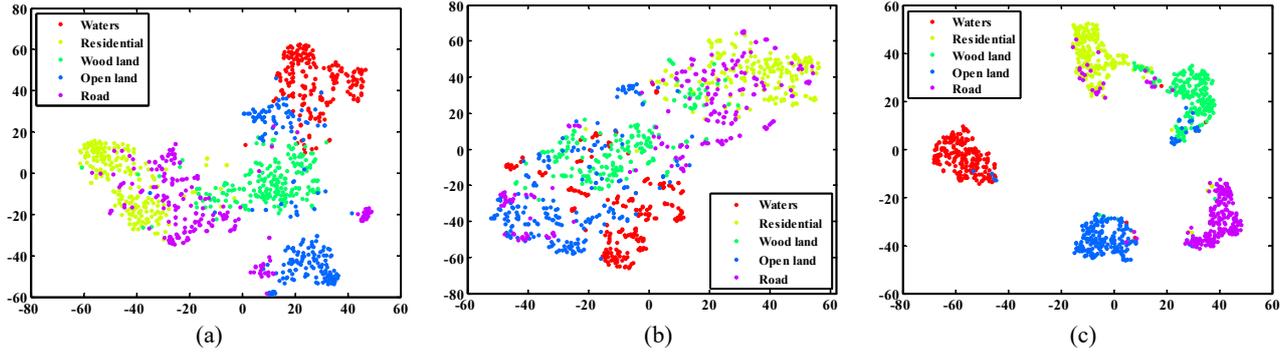

**Fig. 7.** Visualization features of TerraSAR-X data by using t-SNE. (a) spatial features from DSCEN and (b) statistical features from NS-JSM (c) fusional features from Fusion-Net.

### 4.6 Experimental comparisons

To demonstrate the superiority of the proposed method, we compare it with several related methods for HR SAR image classification. These comparison approaches are divided into three groups, including traditional feature extraction, statistical feature extraction, and feature extraction based on deep learning. The details of these comparisons are described as follows.

**Gabor** (Dumitru et al., 2013): Gabor filters are implemented in eight scales and eight directions and the mean of the magnitude of responses is adopted as the feature. A 64-dimensional feature vector was obtained for each SAR image patch.

**CoTF** (Guan et al., 2019): Covariance matrix of the magnitude of Gabor filter responses is calculated. The matrix logarithm operation is applied to map the covariance matrix to the Euclidean space. The upper triangular part of the covariance matrix was used as the feature vector.

**Statistical Dictionary (SD)** (Karine et al. 2017): For a fair comparison, the Gabor filter instead of DT-CWT is implemented on the SAR image. Then, the produced complex subbands magnitudes are modeled by the lognormal model. The obtained statistical parameters are concatenated to build a feature vector for each SAR sample.

**Statistical CNN (SCNN)**: Following the setting of Liu et al. (2020), SCNN contains three convolutional layers, and the feature numbers are 12, 32, and 64, respectively. Then, a global average pooling and a global variance pooling layer are applied to the output of the last convolutional layer to capture statistical features for classification.

**A-ConvNet**: A-ConvNet consists of five convolutional layers, without FC layers being used. A dropout layer with a probability of 0.5 is added after the fourth convolutional layer to prevent overfitting. All parameters are set to the default values as in Chen et al. (2016).

**ResNet** (Fu et al. 2018): An improved Resnet-18 network is used to process SAR images with limited labeled samples. The network architecture contains five residual blocks, and the number of channels is set to 16, 32, 64, 128, and 256 for each layer, respectively. Also, a dropout layer between two stacked convolution layers. This method is used to test the performance of very deep networks in HR SAR image classification.

**MVGG-Net** (Zhang et al., 2021): Following the design architecture of VGG-Net, we migrated the first four convolution blocks of the original VGG, and added a fully connected layer containing 256 neurons to classify SAR images. This method is used to verify the performance of transfer learning and compare it to the proposed DSCEN model on the SAR image classification task.

To ensure the fairness of comparison, the Softmax classifier is adopted to classify the extracted features by all the above algorithms. For the four SAR datasets from different sensors, the comparison method and the proposed method all adopted the same model structure and parameter setting as described in the above section, which can effectively verify the stability and generalization performance of the model.

#### 4.6.1 Results on the TerraSAR-X SAR image

In this section, experiments are conducted on the TerraSAR-X SAR image to compare the classification result with different methods. The compared classification results of AA, OA, and kappa coefficient are reported in Table 9. From Table 9, it can be seen that the proposed Fusion-Net achieves the highest accuracy among the comparison methods, and produces better classification results on most categories. The testing OA, AA and kappa of our approach can reach 90.95%, 90.55%, and 0.8727, respectively. For traditional feature extraction methods, our statistical features based on the NS-JSM have obtained better classification accuracy than Gabor, CoTF, and SD. It indicates that our statistical features can effectively mine the scattering statistics of SAR image patches, making the formed descriptors more robust to various land covers. Compared with the two lightweight SCNN and A-ConvNet, our DSCEN has about a 4% improvement in classification accuracy, which proves that the proposed lightweight DSCEN can learn more effective discriminative feature representation. As for ResNet, we can see that its classification accuracy is the lowest among the deep learning methods. This is due to the limited SAR labeled data that makes it unable to be effectively trained. The classification accuracy of MVGG-Net is close to our DSCEN, which shows that transfer learning can make very deep networks work in SAR classification tasks with limited labeled data. However, the size of MVGG-NET parameters is 37M, which makes it bring more computational burden and occupy more memory. In addition, because the Fusion-Net combines the complementarity of spatial and statistical features, it further improves the classification accuracy. For heterogeneous areas such as residential and texture areas such as woodland and open land, the Fusion-Net has achieved an accuracy improvement of about 2% to 4% compared to the DSCEN that uses a single type of feature classification. In summary, the Fusion-Net shows that the joint consideration of the spatial and statistical features can improve the performance of SAR image classification tasks.

**Table 9**
Classification performance of TerraSAR-X SAR image with different methods.

| Class | Gabor | CoTF | SD | SCNN | A-ConvNet | ResNet | MVGG-Net | NS-JSM | DSCEN | Fusion-Net |
|---|---|---|---|---|---|---|---|---|---|---|
| Waters | 90.86 | 93.01 | 90.77 | **98.85** | 96.61 | 96.54 | 94.42 | 97.21 | 96.41 | 97.63 |
| Residential | 78.95 | 86.06 | 75.78 | 85.97 | 89.36 | 83.51 | 88.51 | 83.93 | 90.94 | **92.69** |
| Woodland | 73.62 | 83.12 | 76.76 | 84.90 | 87.83 | 84.28 | **92.22** | 84.71 | 88.89 | 92.14 |
| Open land | 60.68 | 74.71 | 72.06 | 82.93 | 82.47 | 79.99 | 86.44 | 81.07 | 87.47 | **89.41** |
| Road | 41.27 | 59.47 | 56.45 | 70.56 | 72.88 | 67.62 | 79.24 | 70.80 | 79.89 | **80.87** |
| OA | 70.06 | 80.35 | 74.12 | 84.18 | 86.07 | 82.02 | 88.03 | 82.86 | 89.00 | **90.95** |
| AA | 69.07 | 79.27 | 74.36 | 84.01 | 85.83 | 82.39 | 88.17 | 83.54 | 88.72 | **90.55** |
| $\kappa \times 100$ | 59.91 | 73.02 | 65.17 | 78.14 | 80.67 | 75.36 | 83.34 | 76.43 | 84.62 | **87.27** |

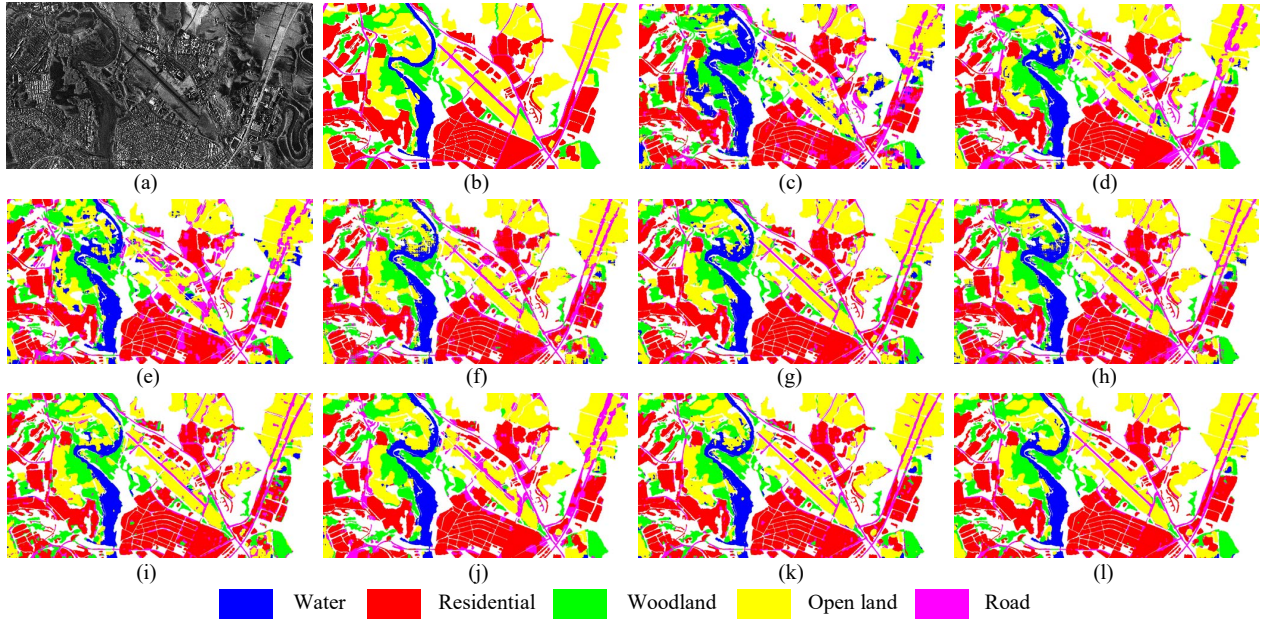

**Fig. 8.** Classification maps of TerraSAR-X SAR image with different methods. (a) Original SAR image. (b) Ground truth. (c) Gabor. (d) CoTF. (e) SD. (f) SCNN. (g) A-ConvNet. (h) ResNet. (i) MVGG-Net. (j) NS-JSM. (k) DSCEN. (l) Fusion-Net.

Fig. 8 depicts the classification result maps of each compared method on the TerraSAR-X image. As is shown in Fig. 8, the Gabor, CoTF, and SD produced serious misclassifications in the road area. Our statistical features show better recognition ability in open land and woodland areas than the other three traditional methods. This implies that our statistical features can suppress the influence of noise or shadows to a certain extent. Moreover, we can see that these deep learning-based methods can roughly identify all categories. But the classification map of our DSCEN has fewer isolated misclassification points, especially in residential and open land areas. Finally, compared with the ground truth, it can be concluded that the proposed Fusion-Net produces the optimal visual effect appearance.

#### 4.6.2 Results on the Gaofen-3 SAR image

Classification results of each approach on the Gaofen-3 data are reported in Table 10. As can be observed, the testing OA, AA, and kappa of our approach are 92.25%, 93.88%, and 0.8991, respectively. The proposed Fusion-Net yields the highest classification accuracies than other approaches, which proves the rationality of the joint consideration of spatial and statistical features. Comparing the Gabor and SD model, we can conclude that statistical modeling is more discriminative than calculating the mean of Gabor filter response. Further, by using our NS-JSM to describe the statistical properties of Gabor wavelet subbands of the SAR image, a better classification performance can be obtained than the other three traditional methods. Compared with SCNN, A-ConvNet, ResNet, and MVGG-Net, the proposed DSCEN performs better than these deep learning methods by 1% in terms of OA. Although the accuracy improvement is relatively small, it indicates the introduction of MSGC and CA blocks in DSCEN can further enhance the classification performance while greatly reducing the number of parameters. Notably, it can be observed that the classification accuracy of our statistical features on the Gaofen-3 SAR data is even better than other deep learning-based methods, which implies that our NS-JSM can perform reasonably well with less training data. An interesting finding is that methods based on second-order statistics such as CoTF, SCNN, and our NS-JSM have achieved more than 90% classification accuracy in woodland, which proves that second-order statistics can help improve the accuracy of objects with complex textures. Besides, it can be seen from the class-specific accuracy that statistical and spatial feature methods have their advantages in the identification of different objects. By using Fusion-Net to fuse the two types of features, it produces the best classification accuracy, which proves that it is necessary to combine statistical and spatial features to process SAR image classification.

Fig. 9 shows the classification result maps of each compared method on the Gaofen-3 SAR image. Traditional classification methods including Gabor and SD contain more misclassified points, especially in mountains and building areas. It can be seen that our statistical features have fewer misclassified pixels in the mountain area than CoTF. The possible reason is that our statistical features also introduce phase information, which is effective for identifying the irregularly textured mountains in Gaofen-3. In addition, our statistical method has better visual effects in buildings and woodland areas than other deep learning methods. On the contrary, our DSCEN method is better at extracting narrow road features and water and open land categories that contain homogeneous areas. Finally, compared with the ground truth, the proposed Fusion-Net can maintain the minimum noise classifications and has a better visual effect, which verifies the effectiveness of multi-feature fusion for improving SAR image classification.

**Table 10**
Classification performance of Gaofen-3 SAR image with different methods.

| Class | Gabor | CoTF | SD | SCNN | A-ConvNet | ResNet | MVGG-Net | NS-JSM | DSCEN | Fusion-Net |
|---|---|---|---|---|---|---|---|---|---|---|
| Mountain | 58.83 | 83.08 | 66.48 | 86.14 | 86.37 | 87.34 | 88.29 | 87.23 | 89.27 | **89.71** |
| Water | 92.73 | 91.40 | 91.96 | 94.61 | 96.44 | 94.59 | 95.15 | 92.67 | **96.45** | 96.36 |
| Building | 67.87 | 88.81 | 73.75 | 81.78 | 78.56 | 80.56 | 82.25 | 86.94 | 82.78 | **89.62** |
| Roads | 81.01 | 93.65 | 86.75 | 93.70 | 94.72 | 95.62 | 96.58 | 94.83 | 96.84 | **98.94** |
| Woodland | 84.61 | 92.16 | 86.12 | 90.43 | 87.61 | 87.24 | 88.47 | **95.51** | 85.84 | 94.37 |
| Open land | 73.42 | 81.14 | 74.81 | 90.25 | 92.65 | 90.29 | 91.29 | 89.52 | **94.56** | 94.29 |
| OA | 74.13 | 87.89 | 78.01 | 87.77 | 87.15 | 87.35 | 88.46 | 89.67 | 89.10 | **92.25** |
| AA | 76.42 | 88.37 | 79.98 | 89.48 | 89.39 | 89.27 | 90.34 | 91.11 | 90.96 | **93.88** |
| $\kappa \times 100$ | 67.41 | 84.22 | 71.99 | 84.21 | 83.39 | 83.66 | 85.06 | 86.61 | 85.85 | **89.91** |

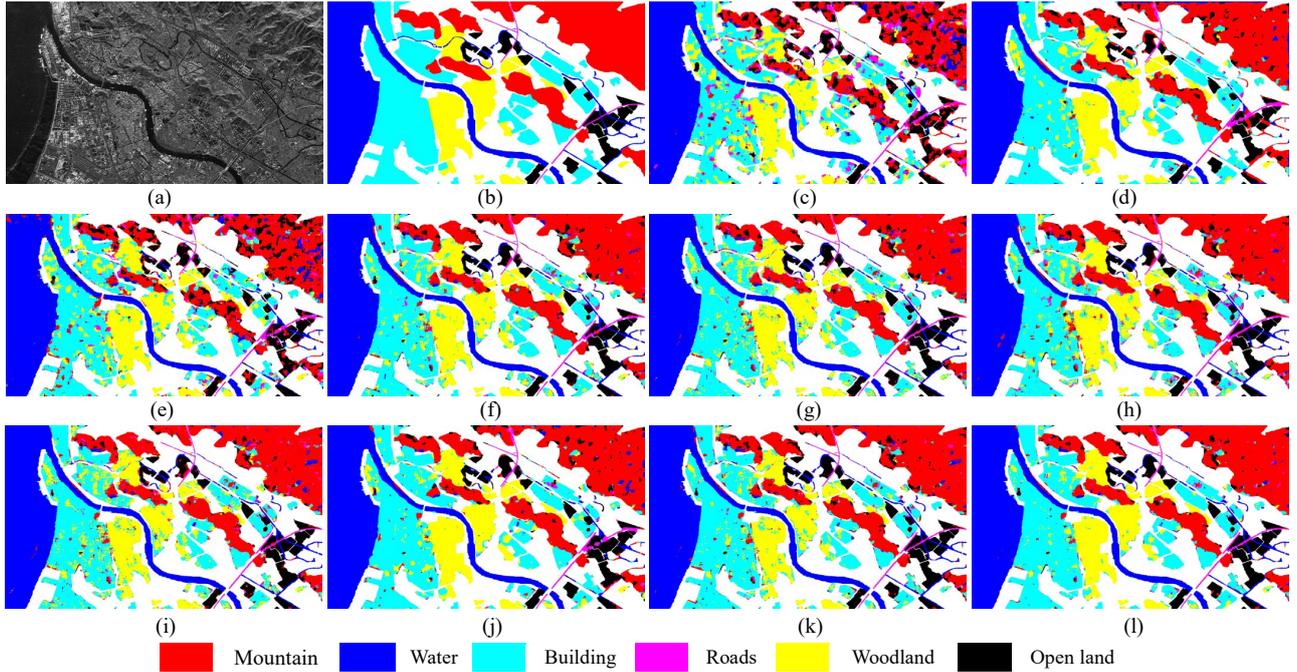

**Fig. 9.** Classification maps of Gaofen-3 SAR image with different methods. (a) Original SAR image. (b) Ground truth. (c) Gabor. (d) CoTF. (e) SD. (f) SCNN. (g) A-ConvNet. (h) ResNet. (i) MVGG-Net. (j) NS-JSM. (k) DSCEN. (l) Fusion-Net.

### 4.6.3 Results on the Airborne SAR image

Classification results of each approach on the Airborne SAR data are shown in Table 11. As is shown in Table 11, the OA, AA, and kappa of our Fusion-Net model are 91.64%, 94.25%, and 0.7417, respectively. It is seen from the compared results that the proposed Fusion-Net achieves the highest classification accuracies. Airborne SAR data contains more categories, but there is a serious class imbalance, where open land occupies most of the pixels in the image. Therefore, it is the most challenging task to classify objects that show a narrow structural appearance and objects that show an extremely complex texture distribution. Compared with Gabor, CoTF, and SD, our statistical features have more than a 3.5% improvement in the kappa coefficient. This indicates that introducing statistical modeling and covariance matrix into the Gabor wavelet subbands of the SAR image can obtain more discriminative statistical features. Secondly, we can see that the deep learning-based methods increase the accuracy of the kappa coefficient by more than 4% compared with other traditional methods. The main reason is that the CNN model has the better representation ability for the narrow structure appearance such as roads, runways, and water areas. Apparently, our DSCEN model achieves the best results of a single type of feature on Airborne SAR data, which demonstrates the DSCEN has better adaptability when dealing with complex shapes and textures in different class objects. Finally, the proposed Fusion-Net can acquire the optimal classification accuracy compared with DSCEN and NS-JSM, which shows that accuracy can be further improved by embedding statistical features into deep spatial features.

Comparisons of classification result maps on the Airborne SAR data are depicted in Fig. 10. As can be seen from Fig. 10, all methods based on traditional feature extraction cannot effectively identify the runway category. In addition, these methods have serious under-classification effects on woodland, roads, and waters. The main reason is that these traditional features pay more attention to extracting the features of SAR image patches from a global perspective, and therefore ignore the expression of local spatial structure relationships. On the contrary, the method based on deep learning has the advantage of learning the spatial local information of the image patches, and it has better feature representation ability in these narrow structural areas and complex urban areas. For the proposed Fusion-Net, it has smoother label consistency on the classification map than other methods, especially in commercial and open land areas. Besides, our Fusion-Net has better classification accuracy at the boundary of the category, so the performance of the visual appearance is further refined.

Table 11
Classification performance of Airborne SAR image with different methods.

| Class | Gabor | CoTF | SD | SCNN | A-ConvNet | ResNet | MVGG-Net | NS-JSM | DSCEN | Fusion-Net |
|---|---|---|---|---|---|---|---|---|---|---|
| Open land | 75.45 | 80.61 | 80.59 | 82.56 | 88.30 | 87.45 | 89.16 | 83.97 | 90.16 | **91.48** |
| Road | 64.73 | 80.00 | 70.38 | 83.51 | 85.14 | 84.30 | 85.73 | 86.35 | 87.85 | **90.68** |
| Water | 78.37 | 97.04 | 93.48 | 93.93 | 98.26 | 98.18 | 98.61 | 97.93 | 98.10 | **98.66** |
| Runway | 95.10 | 96.28 | 95.96 | 98.75 | 99.34 | 99.57 | 98.95 | 99.31 | 98.66 | **99.63** |
| Woodland | 72.70 | 88.00 | 78.44 | 86.63 | 86.85 | 87.25 | 88.00 | 87.40 | **90.17** | 88.51 |
| Residential | 83.74 | **91.78** | 86.16 | 87.35 | 88.28 | 90.61 | 91.77 | 90.71 | 88.84 | 91.56 |
| Commercial | 97.44 | 98.06 | 98.42 | 96.75 | 98.89 | 98.96 | 98.96 | 98.94 | 98.82 | **99.24** |
| OA | 76.65 | 83.68 | 81.62 | 84.61 | 88.86 | 88.53 | 89.82 | 86.11 | 90.77 | **91.64** |
| AA | 81.08 | 90.64 | 86.20 | 90.32 | 92.15 | 92.33 | 93.03 | 92.09 | 93.44 | **94.25** |
| $\kappa \times 100$ | 61.04 | 71.59 | 67.83 | 79.92 | 79.39 | 78.83 | 81.10 | 75.24 | 82.73 | **84.17** |

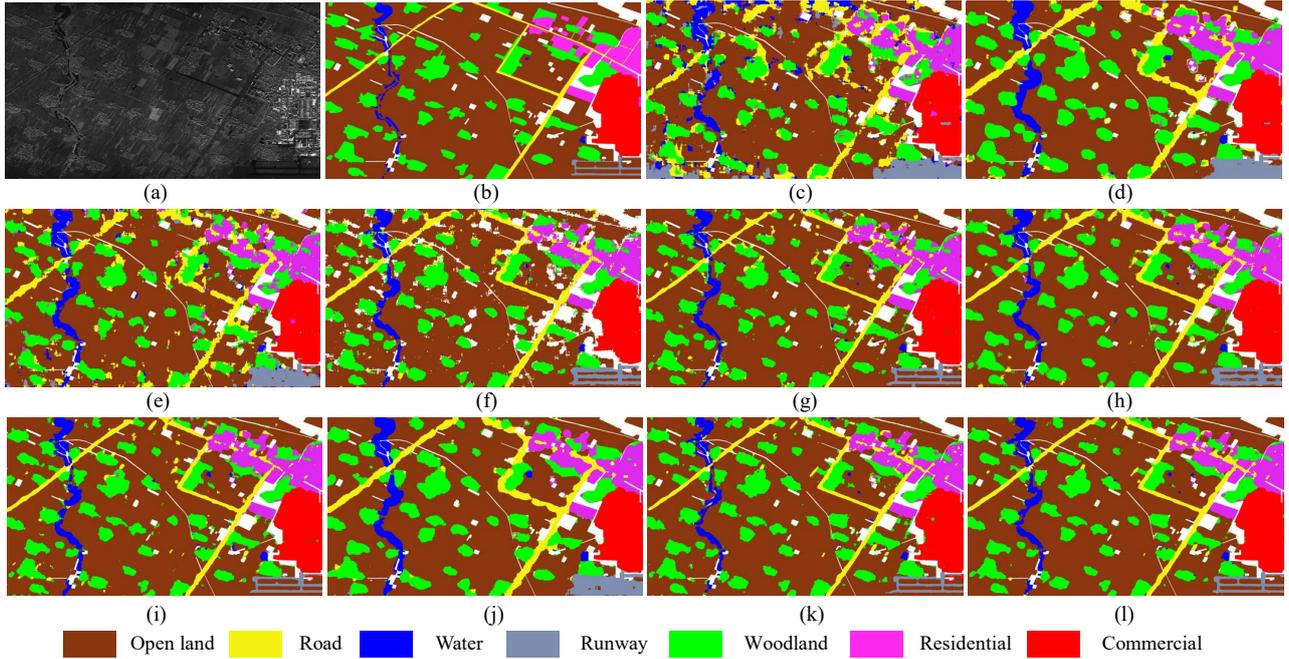

Fig. 10. Classification maps of Airborne SAR image with different methods. (a) Original SAR image. (b) Groundtruth. (c) Gabor. (d) CoTF. (e) SD. (f) SCNN. (g) A-ConvNet. (h) ResNet. (i) MVGG-Net. (j) NS-JSM. (k) DSCEN. (l) Fusion-Net.

4.6.4 Results on the F-SAR image

To illustrate the effectiveness of the proposed Fusion-Net, the experiment was also conducted on the F-SAR image. Table 12 shows the accuracy of per class, OA, AA, kappa coefficient with different methods. Obviously, the open space and vegetation scenes occupy most of the content in the image. These scenes are relatively regular and uniform. Hence, the OA of all comparison methods exceeds 90%. It can be seen that our statistical feature method improves the performance by 1%~2% in accuracy compared with the CoTF and SD. The Gabor method has the lowest classification accuracy in each category due to its insufficient feature discrimination ability. Meanwhile, the deep learning models can effectively identify all categories, but our DSCEN shows relatively better classification performance. This shows that the convolution module we designed can extract more discriminative features. In addition, we found that in residential areas that contain complex structural information, our statistical features have achieved better classification accuracy than DSCEN, which is about a 4% improvement. Finally, by combining complementary information of spatial and statistical features, our Fusion-Net can obtain the best classification performance. An important conclusion is that although the appearance of various land covers in images from different sensors and different resolutions are inconsistent, both spatial and statistical features show their unique advantages in certain specific categories. This also proves that our proposed method is a more potent way to apply it to complex SAR image classification tasks.

Fig. 11 shows the classification result maps by using different methods on the F-SAR image. First, it can be observed that Gabor, CoTF and SD methods have many isolated misclassified pixels in open land and vegetation areas. At the same time, they did not fully detect the residential scenes in some small areas at the bottom of the image. For all deep learning-based methods, they have a small number of misclassified pixels in residential areas. Compared with other models, the proposed Fusion-Net produces the best visual effects, especially where the categories are adjacent, and the boundaries are clearer. Therefore, our proposed Fusion-Net can greatly improve labeling consistency for SAR image classification.

**Table 12**
Classification performance of F-SAR image with different methods

| Class | Gabor | CoTF | SD | SCNN | A-ConvNet | ResNet | MVGG-Net | NS-JSM | DSCEN | Fusion-Net |
|---|---|---|---|---|---|---|---|---|---|---|
| Water | 94.13 | 97.71 | 94.44 | 96.32 | 95.18 | 95.52 | 93.82 | 98.00 | 96.33 | **98.20** |
| Residential | 87.92 | 93.73 | 90.15 | 93.70 | 92.70 | 92.22 | 93.34 | **96.22** | 92.24 | 95.40 |
| Vegetation | 87.75 | 94.84 | 92.01 | 96.08 | 93.50 | 95.14 | 95.39 | 94.58 | 97.55 | **97.78** |
| Open land | 94.02 | 94.13 | 95.60 | 96.75 | 96.34 | 96.29 | 96.91 | 96.72 | **97.99** | 97.80 |
| OA | 89.64 | 94.54 | 92.86 | 96.01 | 94.25 | 95.14 | 95.59 | 95.29 | 97.07 | **97.53** |
| AA | 90.96 | 95.11 | 93.02 | 95.71 | 94.44 | 94.79 | 94.87 | 96.27 | 96.03 | **97.30** |
| $\kappa \times 100$ | 82.19 | 90.34 | 87.57 | 92.97 | 90.03 | 91.48 | 92.26 | 91.71 | 94.77 | **95.59** |

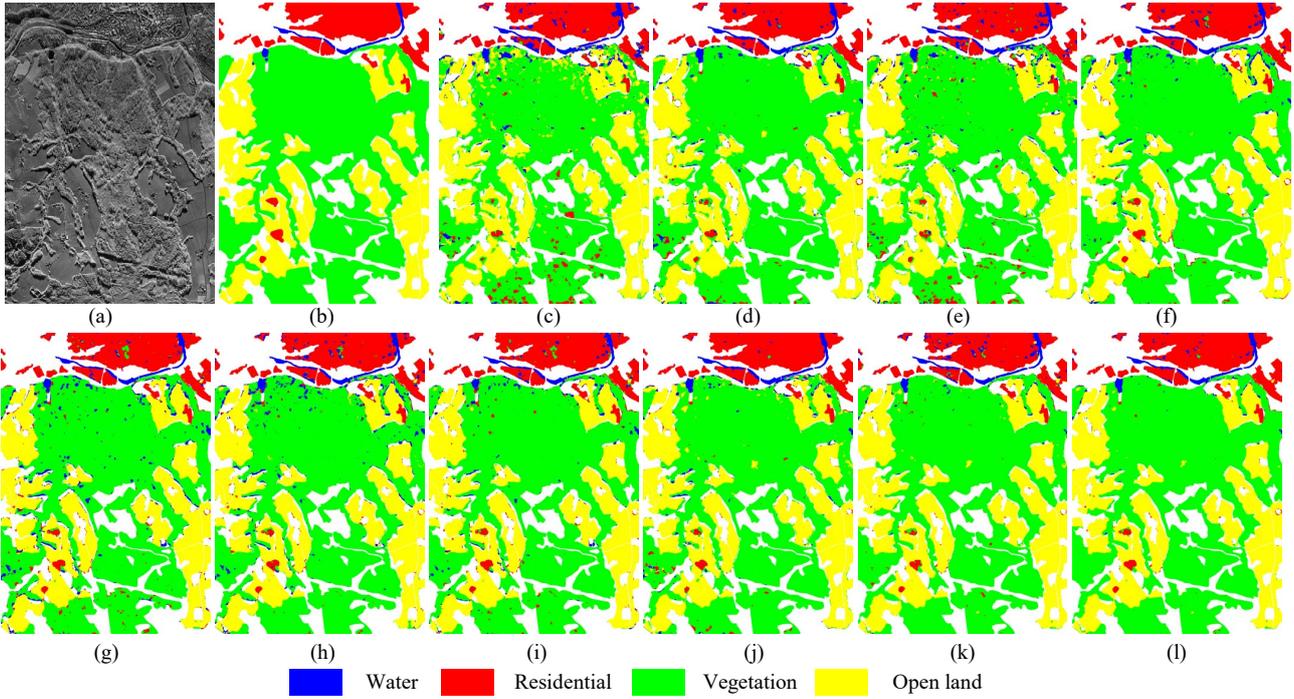

**Fig. 11.** Classification maps of F-SAR image with different methods. (a) Original SAR image. (b) Groundtruth. (c) Gabor. (d) CoTF. (e) SD. (f) SCNN. (g) A-ConvNet. (h) ResNet. (i) MVGG-Net. (j) NS-JSM. (k) DSCEN. (l) Fusion-Net.

## 5. CONCLUSIONS

In this paper, a novel end-to-end Fusion-Net classification model is proposed for HR SAR images, which aims to embed the statistical features into deep spatial features objects in the end-to-end representation learning. In our model, the proposed DSCEN can extract multi-scale spatial features while keeping the fewer model parameters amounts. The NS-JSM can fully mine the statistical properties of the magnitudes and phases of the Gabor wavelet subbands of the SAR image, and form a more compact and robust statistical descriptor. The proposed Fusion-Net can take full advantage of the complementary information of spatial features and statistical features to make the entire classification model achieve a significant accuracy improvement. Experimental results on four SAR images demonstrate that the proposed Fusion-Net yields much higher accuracies and better visual appearance than other related approaches.

In the future, data augmentation such as Mixup (Zhang et al., 2017) or self-supervised learning such as contrastive learning (Chen et al., 2008) will be considered to enhance the feature representation ability of DSCEN. Besides, instead of using the empirical statistical models, we intend to consider adaptively modeling high-order scattering statistics by deep learning to increase the classification capabilities of statistical models.

## DECLARATION OF INTEREST STATEMENT

The authors declare that they have no known competing financial interests or personal relationships that could have appeared to influence the work reported in this paper.

## ACKNOWLEDGEMENTS

This work was supported by the Natural Science Foundation of China (No. 61772390; No. 61871312), the Natural Science Basic Research Plan in Shaanxi Province of China (No. 2019JZ14), and the Civil Space Thirteen Five Years Pre-Research Project (No. D040114).